\pdfoutput=1

\documentclass[11pt]{article}

\usepackage{acl}

\usepackage{times}
\usepackage{latexsym}
\usepackage{subfiles}
\usepackage{graphicx}
\usepackage{multicol}
\usepackage{multirow}
\usepackage[T1]{fontenc}

\usepackage[utf8]{inputenc}

\usepackage{microtype}
\usepackage{amsmath}
\usepackage{amssymb}
\usepackage{subfig}

\usepackage{enumitem}
\usepackage{diagbox}
\usepackage{colortbl}
\usepackage{inconsolata}
\definecolor{blue-violet}{rgb}{0.54, 0.17, 0.89}
\definecolor{bluegray}{rgb}{0.4, 0.6, 0.8}
\definecolor{bleudefrance}{rgb}{0.19, 0.55, 0.91}
\definecolor{darkblue}{rgb}{0.0, 0.0, 0.55}
\definecolor{carnelian}{rgb}{0.7, 0.11, 0.11}
\definecolor{denim}{rgb}{0.08, 0.38, 0.74}
\definecolor{mediumblue}{rgb}{0.0, 0.0, 0.8}
\definecolor{naplesyellow}{rgb}{0.98, 0.85, 0.37}
\definecolor{carminered}{rgb}{1.0, 0.0, 0.22}
\definecolor{lava}{rgb}{0.81, 0.06, 0.13}
	\definecolor{cerisepink}{rgb}{0.93, 0.23, 0.51}
\definecolor{columbiablue}{rgb}{0.61, 0.87, 1.0}
\definecolor{coralred}{rgb}{1.0, 0.25, 0.25}
\definecolor{ao(english)}{rgb}{0.0, 0.5, 0.0}
\definecolor{crimsonglory}{rgb}{0.75, 0.0, 0.2}
\definecolor{lightpastelpurple}{rgb}{0.69, 0.61, 0.85}
\definecolor{navajowhite}{rgb}{1.0, 0.87, 0.68}
\definecolor{tearose(rose)}{rgb}{0.96, 0.76, 0.76}
\definecolor{ruddypink}{rgb}{0.88, 0.56, 0.59}
\definecolor{lightgreen}{rgb}{0.56, 0.93, 0.56}
\definecolor{lightapricot}{rgb}{0.99, 0.84, 0.69}
\definecolor{lime(web)(x11green)}{rgb}{0.0, 1.0, 0.0}
\definecolor{limegreen}{rgb}{0.2, 0.8, 0.2}
\definecolor{magenta(process)}{rgb}{1.0, 0.0, 0.56}
\definecolor{azure(colorwheel)}{rgb}{0.0, 0.5, 1.0}
\definecolor{pastelred}{rgb}{1.0, 0.41, 0.38}
\definecolor{cerise}{rgb}{0.87, 0.19, 0.39}
\definecolor{periwinkle}{rgb}{0.8, 0.8, 1.0}
\definecolor{lightgreen}{rgb}{0.56, 0.93, 0.56}
\definecolor{mediumchampagne}{rgb}{0.95, 0.9, 0.67}
\definecolor{dartmouthgreen}{rgb}{0.05, 0.5, 0.06}
\definecolor{xgreen}{rgb}{0.88, 0.95, 0.83}
\definecolor{airforceblue}{rgb}{0.36, 0.54, 0.66}
\definecolor{uared}{rgb}{0.85, 0.0, 0.3}
\definecolor{lightskyblue}{rgb}{0.53, 0.81, 0.98}
\definecolor{debianred}{rgb}{0.84, 0.04, 0.33}
\definecolor{skyblue}{rgb}{0.53, 0.81, 0.92}
\definecolor{red(ncs)}{rgb}{0.77, 0.01, 0.2}
\definecolor{blush}{rgb}{0.87, 0.36, 0.51}
\definecolor{non-photoblue}{rgb}{0.64, 0.87, 0.93}
\definecolor{lightcornflowerblue}{rgb}{0.6, 0.81, 0.93}
\definecolor{dodgerblue}{rgb}{0.12, 0.56, 1.0}
\definecolor{cadmiumgreen}{rgb}{0.0, 0.42, 0.24}
\definecolor{dollarbill}{rgb}{0.52, 0.73, 0.4}
\definecolor{palegoldenrod}{rgb}{0.93, 0.91, 0.67}
\definecolor{deeppink}{rgb}{1.0, 0.08, 0.58}
\definecolor{lightmauve}{rgb}{0.86, 0.82, 1.0}
\definecolor{darkgoldenrod}{rgb}{0.72, 0.53, 0.04}
\definecolor{deepcerise}{rgb}{0.85, 0.2, 0.53}
\definecolor{darkgreen}{rgb}{0.0, 0.2, 0.13}
 \title{Empowering Language Model with Guided Knowledge Fusion for Biomedical Document Re-ranking}

\author{
 Deepak Gupta and Dina Demner-Fushman  \\
  Lister Hill National Center for Biomedical Communications\\
   National Library of Medicine, National Institutes of Health\\
   Bethesda, MD, USA \\
  {\tt \{firstname.lastname\}@nih.gov
 }\\
  } 

\begin{document}

\maketitle
\begin{abstract}
Pre-trained language models (PLMs) have
proven to be effective for document re-ranking
task. However, they lack the ability to fully 
interpret the semantics of biomedical and health-
care queries and often rely on simplistic pat-
terns for retrieving documents. 
 To address this challenge, we propose an approach that integrates knowledge and the PLMs to guide the model toward effectively capturing information from external sources and retrieving the correct documents.
We performed comprehensive experiments on two
biomedical and open-domain datasets that
show that our approach significantly improves
vanilla PLMs and other existing approaches for
document re-ranking task.

\end{abstract}
\section{Introduction}
Retrieving the relevant information in response to a query involves considering both the explicit constraints indicated in the textual contents of the query as well as implicit knowledge about the domain of interest. Large pre-trained language models (LMs) \cite{devlin2019bert, raffel2020exploring} have become a foundation for most modern information retrieval (IR) systems. While these models have acquired the ability to implicitly encode broad world knowledge and have achieved significant performance on a variety of benchmark tasks, they fall short when provided with examples that are distributionally distinct from those they were fine-tuned on \cite{mccoy2019right}. 

The limitation of LMs is further amplified in the biomedical/clinical setting, where \textbf{(i)} there is a high degree of variability in the form of synonyms and abbreviated words and \textbf{(ii)} the retrieval of relevant information is dependent on focus/intent understanding of the query. In Ex1, from Table-\ref{tab:examples}, both BM25 \cite{robertson2009probabilistic} and MonoT5 \cite{nogueira2020document} models retrieved top documents that include the word ``\textit{CRISPR/Cas9}" from the query. However, the semantics of the query are not considered during retrieval. While the query was about the algorithms for analyzing \textit{``CRISPER/Cas9 knockout screens data''}, both the BM25 (lexical) and MonoT5 missed the document that contains information about 
 \textit{`MaGeCK'}. BM25 retrieved the document that discusses designing \textit{CRISPER/Cas9} based screening experiment for identification of the synthetic lethal target. Similarly, MonoT5 also retrieved the document where \textit{``CRISPER/Cas9 knockout method''} was described in the context of \textit{`Leishmania'}. In the second example (Ex2), the query context is neither explicitly stated in the gold document nor does it contain one salient term (`\textit{chromosome 13}'). It requires domain knowledge to infer that \textit{omodysplasia} is a type of a \textit{``autosomal recessive disorder"} caused by the mutation in a gene on one of the first 22 non-sex chromosomes.
In such cases that require domain knowledge to correctly retrieve the relevant document, both BM25 and MonoT5 fail. 
\begin{table*}[]
\centering
\resizebox{\textwidth}{!}{%
\begin{tabular}{l|l|l|l}
\hline
\multicolumn{1}{c|}{\textbf{Question}} &
  \multicolumn{1}{c|}{\textbf{\begin{tabular}[c]{@{}c@{}}Top Retrieved Document\\ (BM25)\end{tabular}}} &
  \multicolumn{1}{c|}{\textbf{\begin{tabular}[c]{@{}c@{}}Top Retrieved Document\\ (MonoT5)\end{tabular}}} &
  \multicolumn{1}{c}{\textbf{Gold Document}} \\ \hline \hline
\begin{tabular}[c]{@{}l@{}}\textbf{Ex1: }Which algorithms have been developed \\ for analysing \textcolor{azure(colorwheel)}{CRISPR/Cas9 knockout} \\ \textcolor{azure(colorwheel)}{screens} data?\end{tabular} &
  \begin{tabular}[c]{@{}l@{}} \textcolor{azure(colorwheel)}{CRISPR/Cas9}, an RNA guided endonuclease system \\ is the most recent technology for this work. Here, we \\ have discussed the major considerations involved in \\ designing a \textcolor{azure(colorwheel)}{CRISPR/Cas9} based \textcolor{azure(colorwheel)}{screening} experiment \\ for identification of synthetic lethal targets.\end{tabular} &
  \begin{tabular}[c]{@{}l@{}}We describe here in detail a simple, rapid, and \\ scalable method for \textcolor{magenta(process)}{CRISPR-Cas9-mediated} \\ gene  \textcolor{azure(colorwheel)}{knockout} and tagging in Leishmania. This\\ method details how to use simple PCR to generate \\ (1) templates for single guide RNA (sgRNA) \\ transcription in cells expressing  \textcolor{magenta(process)}{Cas9} and T7 \\ RNA polymerase..\end{tabular} &
  \begin{tabular}[c]{@{}l@{}}We propose the Model-based Analysis \\ of Genome-wide \textcolor{azure(colorwheel)}{CRISPR/Cas9 Knockout}\\ (\textcolor{limegreen}{MAGeCK}) method for prioritizing \\ single-guide RNAs, genes and pathways \\ in genome-scale \textcolor{azure(colorwheel)}{CRISPR/Cas9 knockout} \\ screens.\end{tabular} \\ \hline
\begin{tabular}[c]{@{}l@{}}\textbf{Ex2: }What rare disease is associated with a \\ mutation in the \textcolor{azure(colorwheel)}{GPC6 gene} on \\ \textcolor{azure(colorwheel)}{chromosome 13}?\end{tabular} &
  \begin{tabular}[c]{@{}l@{}}..We report the construction of a high-resolution 4 Mb\\  sequence-ready BAC/PAC contig of the GPC5/\textcolor{azure(colorwheel)}{GPC6}\\ gene cluster on  \textcolor{azure(colorwheel)}{chromosome} region  \textcolor{magenta(process)}{13q32}.\end{tabular} &
  \begin{tabular}[c]{@{}l@{}}The human \textcolor{magenta(process)}{gamma-sarcoglycan gene} was \\ mapped to \textcolor{magenta(process)}{chromosome 13q12}, and deletions\\ that alter its reading frame were identified in \\ three families and one of four sporadic cases\\ of SCARMD.\end{tabular} &
  \begin{tabular}[c]{@{}l@{}}.. The proband had normal molecular analysis \\ of the \textcolor{magenta(process)}{glypican 6 gene} (\textcolor{azure(colorwheel)}{GPC6}), which was \\ recently reported as a candidate for \textcolor{limegreen}{autosomal} \\ \textcolor{limegreen}{recessive omodysplasia}.  Mild rhizomelic \\ shortening of the lower extremities has not been\\ previously reported...\end{tabular} \\ \hline
\end{tabular}
}
\caption{Sample questions and gold document from the BioASQ dataset along with the top retrieved documents using BM25 and MonoT5 methods. Lexical and semantic matches considering context are shown in \textcolor{azure(colorwheel)}{blue} and \textcolor{magenta(process)}{pink}. The highlighted texts in \textcolor{limegreen}{green} represent the requirements of domain knowledge to retrieve the correct document. }
\label{tab:examples}
\end{table*}
These findings highlight that LMs lack semantic interpretation of queries and oftentimes depend on naïve patterns to retrieve information rather than using more structured reasoning that effectively amalgamates information provided in the context with external knowledge. In the past, there have been numerous research efforts to effectively fuse domain knowledge in LMs, which has been observed to be beneficial in capturing semantic context in various NLP tasks \cite{ghazvininejad2018knowledge,huang2020knowledge,yasunaga2021qa}, yet, so far to the best of our knowledge, there has been no exploration towards integrating external knowledge in neural IR, both in open domain and much-needed biomedical/clinical domain.

To address the aforementioned issues, in this work, we propose {GraphMonoT5}, an effective approach that fuses the external knowledge into the pre-trained language model for the document retrieval task. {GraphMonoT5} takes the query, document, and graph as input and learns to predict the relevant score for the document against the query. The proposed GraphMonoT5 is built upon the encoder-decoder T5 model, and the T5 encoder layer is complemented with the graph neural network (GNN). The former takes query and document as input and later is used to reason over the underlying knowledge graph (KG) with entities as nodes and relationships between them as edges. With the use of mutual information based bottleneck interaction representations, we develop a strategy to effectively fuse the language and graph representation and allow a two-way exchange of information between the two modalities: text and graph. The representations of text and graph are generated via the T5 encoder and the GNN, respectively. 
 
The extensive experiments on biomedical and open-domain datasets show that {GraphMonoT5} achieves better performance compared to the existing re-ranking approaches.

\section{Related Works}
The cross-attention based neural re-ranking methods \cite{han2020learning,nogueira2020document,chen2021co} take the output of a first-stage retrieval system, such as BM25, and reorder the retrieved documents to push more relevant documents to the top of the retrieval results. There have been studies \cite{hui2022ed2lm,ju2021text,sachan2022improving,ma2021zero} that focus on minimizing the computational overhead of cross-attention models, and they designed new objective functions and the scoring mechanisms that can achieve comparable performance to cross-attention models. The BioASQ (Large-scale biomedical semantic indexing and question answering) shared task enables research in biomedical document retrieval \cite{tsatsaronis2015overview}. However, most of the systems proposed for the biomedical document retrieval task have primarily relied upon term-matching algorithms. Some of the recent systems have made progress by leveraging neural re-ranking of retrieved candidates \cite{pappas2020aueb,almeida2020bit,brokos2018aueb,lu2022zero}.
 Recently, \citet{luo2022improving} proposed Poly-DPR and two new pre-training tasks to overcome the limitations of neural retrieval models for the biomedical domain. In contrast, our study focuses on integrating external knowledge into PLMs with a specially designed modalities fusion strategy that helps in improving the ranking performance.

\section{Methodology}\label{sec:method}

\subsection{Background}
\label{monot5-background}
Our proposed re-ranking approach {GraphMonoT5} is based on the MonoT5 model that utilizes the encoder-decoder based T5 \cite{raffel2020exploring} model to calculate a relevance score that provides a quantitative indication of the degree to which a candidate document $d$ is pertinent to a query $q$. The input prompt to the MonoT5 model is: 
\begin{equation}
\label{t5-input}
     \text{Query:~ [$q$] ~ Document: } ~ [d] ~ \text{Relevant: }
 \end{equation}
The MonoT5 model is fine-tuned to generate the words ``true'' for relevant documents or ``false'' for the documents non-relevant to the query. 

During inference, the candidate documents are re-ranked based on the probability of the ``true'' token.
\subsection{Proposed Model}
Our proposed {GraphMonoT5} model is the result of the augmentation of the PLM with the graph reasoning modules over KG for effectively re-ranking the candidate documents against the query. We describe the KG construction and KG-enriched ranking in the following subsections:
\subsubsection{Knowledge Graph Construction}
The knowledge graph is a multi-relational graph $\mathcal{G}=(\mathcal{V}, \mathcal{E})$ with entity nodes $\mathcal{V}$ and edges $\mathcal{E} \subseteq \mathcal{V} \times \mathcal{R} \times \mathcal{V}$ that connect nodes in $\mathcal{V}$ with the set of relations $\mathcal{R}$. Given a query-document pair ($q, d$), following the work of \citet{lin-etal-2019-kagnet}, we link the entities mentioned in the question and document to the KG $\mathcal{G}$. The nodes corresponding to query $q$ and document $d$ are denoted by $\mathcal{V}_q \subseteq \mathcal{V}$ and $\mathcal{V}_d \subseteq \mathcal{V}$, respectively. The total of nodes of the query-document pair is denoted by $\mathcal{V}_{q,d} = \mathcal{V}_q \cup \mathcal{V}_d$. Since the KG $\mathcal{G}$ can include millions of nodes and edges; therefore a subgraph $\mathcal{G}_{q,d}= (\mathcal{V}_{q,d}, \mathcal{E}_{q,d})$ of the KG $\mathcal{G}$ which contains all the nodes on the 2-hop paths between nodes in $\mathcal{V}_{q,d}$ is considered for the query-document pair. 
\subsubsection{KG-enriched Seq2Seq Ranking} Our KG-enriched seq2seq ranking approach consists of \textbf{(a)} $R$ layers \texttt{T5-encoder} model to encode the language context, \textbf{(b)} graph neural network to model the subgraph of the query-document pair, \textbf{(c)} $S$ layers language-graph interaction component to fuse the language and graph representations, and \textbf{(d)} T5-decoder model to predict the query-document relevance score. Following, \citet{zhang2021greaselm}, we use an interaction token $t_{int}$ and interaction node $n_{int}$ to pass the information across the language and graph modalities. The interaction token $t_{int}$ is prepended to the token sequence $\{t_1, t_2,\ldots, t_N\}$ of query-document pair ($q,d$) (\textit{cf.} Eq. \ref{t5-input}) and $n_{int}$ is connected to $n_{int}$ node that is linked to all the nodes in  $\mathcal{G}_{q,d}$.
\paragraph{Language Representation:} Given the token sequence $\mathcal{T}=\{t_{int}, t_1, t_2,\ldots, t_N\}$, first we pass the sequence $\mathcal{T}$ into the first layer of the \texttt{T5-encoder} to obtain the hidden state representations $H^1= \{h_{int}^1, h_1^1, h_2^1,\ldots, h_N^1\} \in \mathcal{R}^{{(N+1)} \times d_l}$. Hidden state representation $H^{l}$ at $l^{th}$ layer is passed to the $(l+1)^{th}$ layer of \texttt{T5-encoder} \cite{raffel2020exploring} to encode and obtain the representation $H^{l+1}$. Following this, we extracted the representation from \texttt{T5-encoder} for $l=1,2,\ldots,R$:
\begin{equation}
\footnotesize
    \label{lang_representation}
    h_{int}^{l+1}, h_1^{l+1}, \ldots, h_N^{l+1} = \texttt{\texttt{T5-encoder}}(h_{int}^{l}, h_1^{l}, \ldots, h_N^{l})
\end{equation}
To fuse the language and graph representation, we also extracted the hidden state representation from an additional $S$ layers of \texttt{\texttt{T5-encoder}}; however, at layer $l$ the interaction token representation $h_{int}^{l}$ is fused with the interaction node representation (to be discussed shortly) to amalgamate the knowledge feature with the language model feature.
\paragraph{Graph Representation:} Given the question-document pair sub-graph $\mathcal{G}_{q,d}= (\mathcal{V}_{q,d}, \mathcal{E}_{q,d})$ with nodes $\{n_{int}, n_1, n_2,\ldots, n_M\}$, we first compute the node embeddings (details in \textbf{Appendix}) using the pre-trained knowledge graph embeddings $U^1= \{u_{int}^1, u_1^1, u_2^1,\ldots, u_M^1\} \in \mathcal{R}^{{(M+1)} \times d_g}$. We utilized the graph neural network discussed in \citet{zhang2021greaselm,velivckovic2018graph} to compute the node representation by propagating the information across the nodes in the subgraph $\mathcal{G}_{q,d}$.
The subgraph node representation $U^{l}$ at $l^{th}$ layer of GNN is passed to the $(l+1)^{th}$ layer of GNN to encode and obtain the representation $U^{l+1}$. Following this, we extracted the representation from GNN for $l=1,2,\ldots, S$:
\begin{equation}
\small
    \label{graph_representation}
    u_{int}^{l+1}, u_1^{l+1}, \ldots, u_M^{l+1} = \texttt{GNN}(u_{int}^{l}, u_1^{l}, \ldots, u_M^{l})
\end{equation}

\paragraph{Language-graph Interaction:} On a given layer $l \in S$, we aim to effectively fuse the modalities by using the interaction token representation $h_{int}^{l}$ and interaction node representation $u_{int}^{l}$. Towards this, first, we obtained the fused representation $x^l=f(h_{int}^{l}\oplus u_{int}^{l})$ with a two-layer feed-forward network $f$. The fused representation $x^l$ may contain redundant information. To overcome this issue, we introduce mutual information (MI) based feature fusion which aims to minimize the MI $\mathcal{I}(x^l;z^l)$ between the compressed encoded representation $z^l$ and the concatenated representation $x^l$. Formally given two random variables $x^l$ and $z^l$, their MI is defined as follows:
\begin{equation}
\small
\label{mi-fusion}
    \begin{split}
        \mathcal{I}(x^l;z^l) &= D_{KL}(p(x^l,z^l) || p(x^l)p(z^l))\\
       & \leq  \alpha \mathbb{E}_{z^l \sim p(z^l|x^l)} [ D_{KL}(p(z^l|x^l)|| q(z^l) )] \\
        & \leq  \alpha M(x^l;z^l) 
    \end{split}
\end{equation}
where, $\alpha$ is a constant and $D_{KL}$ denotes the KL divergence (proof in \textbf{Appendix}). We model the $p(z^l|x^l)$ using a parameterized Gaussian distribution $\mathcal{N}(\mu_z^l, \Sigma_z^l)$ with mean $\mu_z^l$ and variance $\Sigma_z^l$.
To compute the gradients through random variables, we follow the reparametrization trick \cite{kingma2013auto} with standard normal distribution $\epsilon \sim \mathcal{N}(0, I)$ to calculate $z^l= \mu_z^l+ \Sigma_z^l\epsilon$. Later, we split $z^l$ into the $\tilde{h}_{int}^l$ and $\tilde{u}_{int}^l$ for further computation of the token and node, respectively. With the virtue of Transformer network \cite{vaswani2017attention} and GNN, the fused representation is mixed with the remaining tokens and nodes of the subgraph. 
The graph-augmented representations from the KG-enriched \texttt{T5-encoder} are passed to the \texttt{T5-decoder} to predict the query-document relevance score as discussed in Section \ref{monot5-background}.
\paragraph{Training and Inference}:
The network is trained by maximizing the log-likelihood of the document given the query and minimizing the mutual information on each layer of the KG-enriched \texttt{T5-encoder}. Formally,
\vspace{-0.8em}
\begin{equation}
    J=p(y|\mathcal{T}) - \frac{\alpha}{S}\sum_{l=1}^{S}  M(x^l;z^l)
\end{equation}

where $y \in \{\text{`true'}, \text{`false'}\}$ is the predicted token from T5 model given the input token sequence $\mathcal{T}$. We use Monte Carlo sampling \cite{shapiro2003monte} to compute the approximated value of the mutual information. 
\begin{table}[]
\centering
\resizebox{\linewidth}{!}{%
\begin{tabular}{l|cc|cc|cc}
\hline
\multicolumn{1}{c|}{\multirow{2}{*}{\textbf{Models}}} & \multicolumn{2}{c|}{\textbf{BioASQ8B}}  &\multicolumn{2}{c|}{\textbf{TREC-COVID}}     & \multicolumn{2}{c}{\textbf{HotPotQA}}    \\ \cline{2-7} 
\multicolumn{1}{c|}{}                                & \multicolumn{1}{c}{\textbf{R@100}} & \textbf{nDCG@10}  & \multicolumn{1}{c}{\textbf{R@100}} & \textbf{nDCG@10} & \multicolumn{1}{c}{\textbf{R@100}} & \textbf{nDCG@10} \\ \hline \hline

\textbf{DeepCT} \cite{dai2020context}      &$0.699$ & $0.407$ & $0.347$ &  $0.406$ & $0.731$ & $0.503$ \\ 
\textbf{SPARTA} \cite{zhao2021sparta}      & $0.351$ & $0.351$ & $0.409$&  $0.538$ &  $0.651$ & $0.492$ \\ 

\textbf{DPR} \cite{karpukhin2020dense}         & $0.256$ & $0.127$ & $0.212$ &  $0.332$ & $0.591$ & $0.391$ \\ 
\textbf{ANCE} \cite{xiong2020approximate}       & $0.463$ & $0.306$ &  $0.457$ &   $0.654$ & $0.578$ & $0.456$ \\ 
\textbf{TAS-B} \cite{hofstatter2021efficiently}       & $0.579$ & $0.383$ & $0.387$&   $0.481$&$0.728$ & $0.584$ \\ 
\textbf{GenQ} \cite{thakur2021beir}        &$0.627$ & $0.398$ & $0.456$ &  $0.619$ & $0.673$ & $0.534$ \\ 

\begin{tabular}[c]{@{}l@{}}\textbf{ColBERT}\\  \cite{khattab2020colbert} \end{tabular}   & $0.645$ & $0.474$ &$0.464$ &  $0.677$ &   $0.748$ & $0.593$ \\ 
\hline

\textbf{BM25} \cite{robertson2009probabilistic}        & $0.745$&$0.488$ &  $0.508$ & $0.688$     &$0.763$& $0.602$    \\ 
\textbf{MonoT5} \cite{nogueira2020document}      &  $0.745$  &    $0.489$   &  $0.508$&  $0.685$&  $0.763$  &  $0.648$     \\  \hline
\textbf{Proposed (GraphMonoT5)} &  ${0.745}$   & $\textbf{0.520}$    &  ${0.508}$&  $\textbf{0.701}$  &   ${0.763}$   &   $\textbf{0.667}$    \\ 
      \quad w/o \textbf{ MI Fusion}     &   $0.745$    & $0.499$  &  $0.508$&  $0.683$     &  $0.763$  &$0.637$       \\ \hline
      
\end{tabular}%
}
\caption{Performance comparison of our proposed method with the existing approaches on respective datasets. R@$100$ refers to the Recall@$100$.}
\label{tab:main-results}
\end{table}

\section{Results and Analysis}
\label{sec:res-analysis}

\paragraph{Datasets and Knowledge Sources:}
We evaluated our proposed {GraphMonoT5} model on two biomedical BioASQ8B \cite{nentidis2020overview}, TREC-COVID \cite{voorhees2021trec} and one open domain HotPotQA \cite{yang2018hotpotqa} datasets. We utilized \textit{ConceptNet} \cite{speer2017conceptnet} to extract the knowledge for the HotPotQA dataset, and biomedical knowledge graph from the Unified Medical Language System (UMLS) \cite{bodenreider2004unified} and DrugBank \cite{wishart2018drugbank} knowledge sources for the BioASQ8B dataset. The detailed statistics of the datasets, knowledge sources, and implementation details are given in the \textbf{Appendix}.

\begin{figure}%
 \vspace{-1em}
    \centering
    \subfloat[\centering BioASQ8B]{{\includegraphics[width=0.47\linewidth]{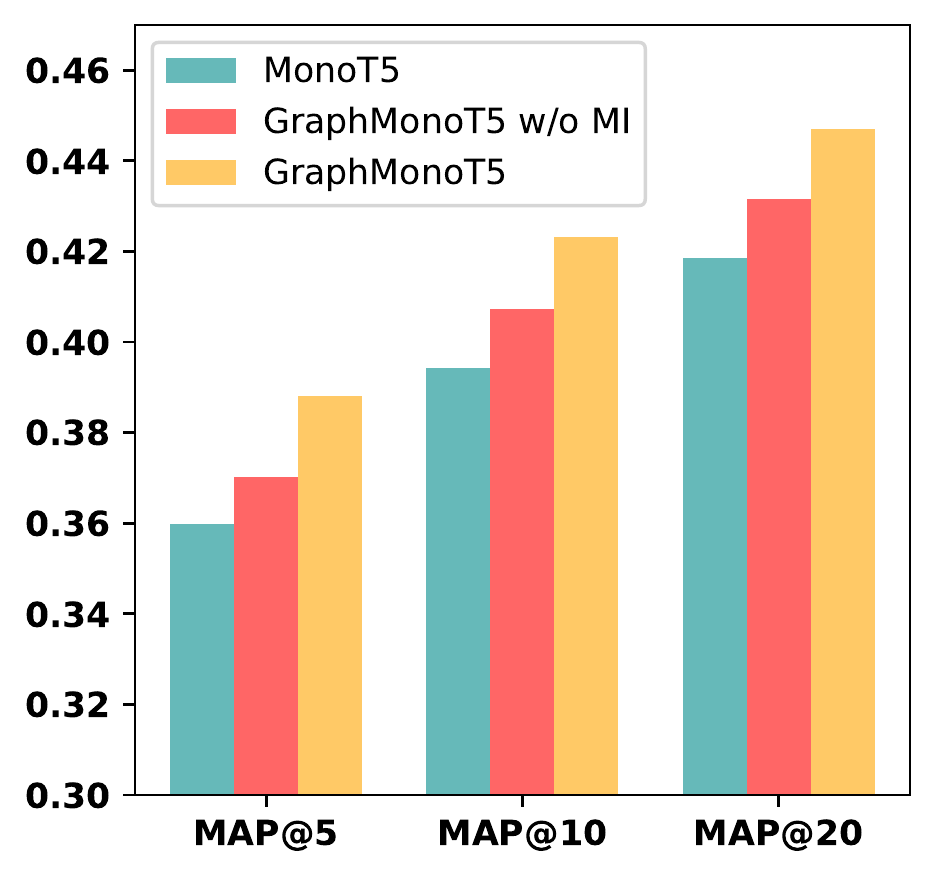} }}%
    \hspace{0.1em}
    \subfloat[\centering HotPotQA]{{\includegraphics[width=0.47\linewidth]{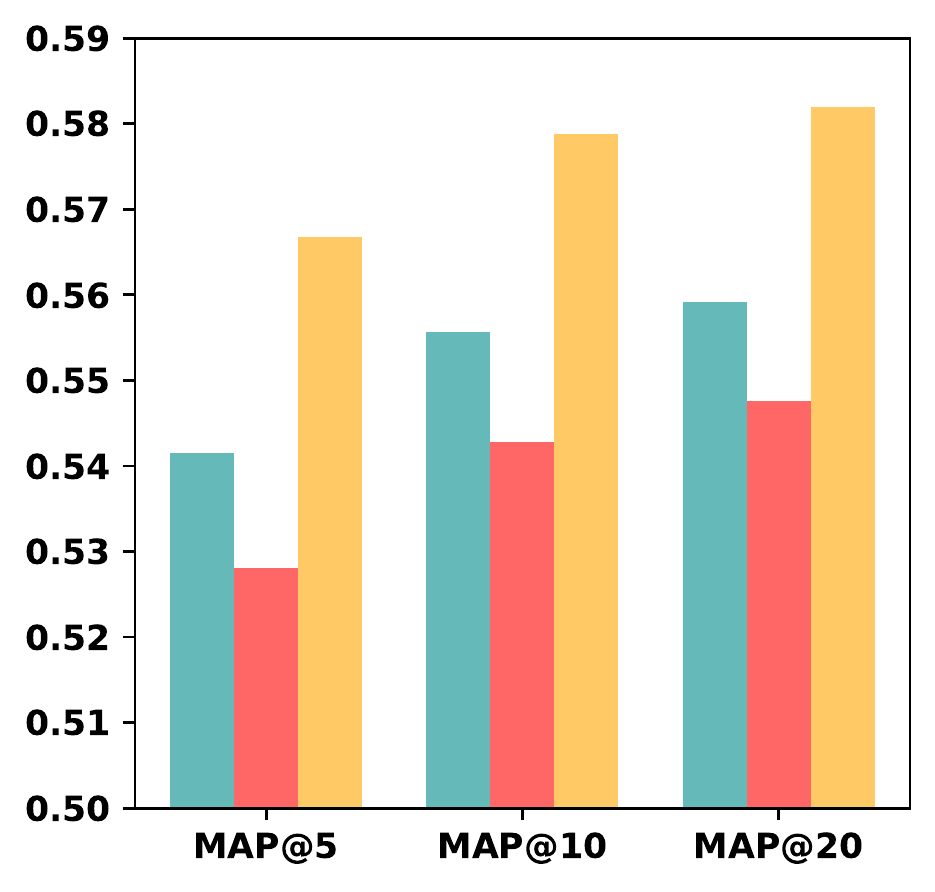} }}%
    \caption{Performance comparison of models in terms of MAP$@k$ for BioASQ8B and HotPotQA test datasets.}%
    \label{fig:results-map}%
\end{figure}
\begin{table}[t]
\resizebox{\linewidth}{!}{%
\begin{tabular}{l|cccccc}
\hline
          \textbf{Methods}        & \textbf{B1}     & \textbf{B2}     & \textbf{B3}     & \textbf{B4}     & \textbf{B5}     & \textbf{Mean}    \\ \hline \hline
\citet{kazaryan2020transformer}
   & $0.3346$ & $0.3304$ & $0.4351$ & $0.3600$& $\textbf{0.4825}$ & $0.3885$ \\
\citet{pappas2020aueb}
   & $0.3359$ & $0.3181$ & $0.4510$ & $0.4163$ & $0.4657$ & $0.3974$\\
\citet{luo2022improving} & $0.3002$ & $0.3131$ & $0.3979$ & $0.4218$ & $0.3799$ & $0.3626$ \\ \hline
\textbf{Proposed (GraphMonoT5)}  & $\textbf{0.3906}$ &	$\textbf{0.3943}$	& $\textbf{0.4697}$	& $\textbf{0.5190}$ &	$0.4168$ &	$\textbf{0.4308}$ \\ 
\hline 

\hline
\end{tabular}%
}
\caption{Comparison of the proposed method with the state-of-the-art approaches on BioASQ8B test batches in terms of MAP score.}
\label{tab:bioasq-sota-results}
\end{table}
\paragraph{Results:} We have presented the results in Table \ref{tab:main-results}, which demonstrates that the GraphMonoT5 model equipped with knowledge-graph outperforms the existing approaches on BioASQ8B, TREC-COVID, and HotPotQA test datasets. Since the TREC-COVID dataset does not contain the training set, therefore, we evaluated the model trained on the BioASQ8B dataset on the test set of TREC-COVID in a zero-shot setting. With GraphMonoT5, we observed an improvement of $3.1$, $1.6$, and $1.9$ nDCG@10 points over the vanilla MonoT5 model on BioASQ8B, TREC-COVID, and HotPotQA datasets, respectively. Furthermore, compared to BM25, we observed an improvement of $3.2$, $1.3$, and $6.5$ nDCG@10 points on respective datasets.
We have also provided a performance comparison of our proposed approach with the best systems of the BioASQ8 challenge and recent work of \citet{luo2022improving} in Table \ref{tab:bioasq-sota-results}. The results allow for two important claims \textbf{(1)} knowledge-enriched PLMs help to re-rank the documents more accurately compared to the vanilla PLMs and \textbf{(2)} mutual information based knowledge-fusion is an appropriate strategy to fuse the language and graph information.\\
\textbf{Analysis:} To analyze the role of mutual information based objective function, we trained the model with only cross-entropy loss and observed the decrements of $2.1$, $1.8$, and $3.0$ nDCG@10 points on BioASQ8B, TREC-COVID, and HotPotQA dataset respectively. We have also provided (\textit{cf.} Fig \ref{fig:results-map}) the comparison of the approaches in terms of MAP, which shows that the GraphMonoT5 method with mutual information fusion outperforms the MonoT5 and concatenation based fusion on BioASQ8B and HotPotQA datasets.
\section{Conclusion}
In this work, we proposed an effective approach to re-rank the documents by utilizing the knowledge graph and integrating the external knowledge into the PLMs. To effectively fuse the language and graph information in the knowledge-enriched framework, we introduced a mutual information-based objective function, which ensures the fused representations are non-redundant and informative in nature. Extensive experiments on biomedical and open-domain datasets show the effectiveness of the proposed approach.

\bibliographystyle{acl_natbib}
\bibliography{anthology,custom}
\appendix
\section{Language-graph Interaction}\label{sec:lang-graph-interaction}
Formally given two random variables $x$ and $z$, their MI is defined as follows:
\begin{equation}
\small
\label{mi-fusion-appendix}
    \begin{split}
        \mathcal{I}(x;z) &= D_{KL}(p(x,z) || p(x)p(z))\\
        &=\int  p(x,z) \log\frac{p(x,z)}{p(x)p(z)} dxdz \\
          &=\int p(x,z) \log\frac{p(z|x)}{p(z)} dxdz\\
           &=\int p(x,z) \log p(z|x)dxdz - \int p(z) \log p(z)dz
    \end{split}
\end{equation}
We know that KL divergence follows the property that $D_{KL}(p(z)||q(z)) \geq 0$; where $q(z)$ is a variational approximation to the distribution $p(z)$, therefore, $\int p(z)\log p(z)dz \geq \int p(z)\log q(z)dz$. Following this, we can rewrite Eq. \ref{mi-fusion-appendix} as follows: 
\begin{equation}
\small
\label{mi-fusion-with-property}
    \begin{split}
        \mathcal{I}(x;z)
           &=\int p(x,z) \log p(z|x)dxdz - \int p(z) \log p(z) dz \\
       & \leq   \int p(x,z) \log p(z|x) dxdz - \int p(z) \log q(z)dz \\
         & \leq   \int p(x)p(z|x) \log \frac{p(z|x)}{q(z)} dxdz  \\
          & \leq  \alpha \mathbb{E}_{z \sim p(z|x)} [ D_{KL}(p(z|x)|| q(z) )]  \\
    \end{split}
\end{equation}

\section{Datasets and Knowledge Sources}
We evaluated our proposed \textsc{GraphMonoT5} model on two biomedical and one open domain datasets. For biomedical domains, we train the model on the training collection of the BioASQ8B \cite{nentidis2020overview} dataset, the network hyper-parameters are tuned on a batch four test collection of BioASQ7B, and performance is reported on the five different test collections (B1, B2, B3, B4, and B5) each of $100$ queries of BioASQ8B and TREC-COVID \cite{voorhees2021trec} dataset. To report the performance of the proposed approach on the open domain, we considered HotPotQA \cite{yang2018hotpotqa} dataset. The PubMed and Wikipedia corpus from \citet{thakur2021beir} are considered to retrieve the relevant documents. We utilized \textit{ConceptNet} \cite{speer2017conceptnet}, an open-domain knowledge graph, to extract the knowledge for the HotPotQA dataset and biomedical knowledge graph from \citet{zhang2021greaselm} that is developed by integrating the Unified Medical Language System (UMLS) \cite{bodenreider2004unified} and DrugBank \cite{wishart2018drugbank} knowledge sources to extract the knowledge from BioASQ datasets.
The detailed statistics of the datasets and knowledge graph are shown in Table \ref{tab:dataset}.
\begin{table}[]
\resizebox{\linewidth}{!}{%
\begin{tabular}{l|c|c|c|c|c|c}
\hline
 \textbf{Datasets} &
  \textbf{\begin{tabular}[c]{@{}c@{}}Training\\ Query-doc \\ Pairs\end{tabular}} &
  \textbf{\begin{tabular}[c]{@{}c@{}}Dev\\ Queries\end{tabular}} &
  \textbf{\begin{tabular}[c]{@{}c@{}}Test\\ Queries\end{tabular}} &
  \multicolumn{1}{l|}{\textbf{Corpus}} &
  \textbf{\begin{tabular}[c]{@{}c@{}}KG\\ Nodes\end{tabular}} &
  \textbf{\begin{tabular}[c]{@{}c@{}}KG\\ Edges\end{tabular}} \\ \hline \hline
\textbf{BioASQ8B}   & 32,916  & 100   & 500   & 14,914,602 & 9,958   & 44,561    \\ 
\textbf{TREC-COVID} & -       & -     & 50    & 171,332    & -       & -         \\ 
\textbf{HotPotQA}   & 170,000 & 5,447 & 7,405 & 5,233,329  & 799,273 & 2,487,810 \\ \hline \hline
\end{tabular}%
}
\caption{Statistics of the datasets used in the experiments.}
\label{tab:dataset}
\end{table}
\section{Implementation \& Training Details:}
\paragraph{Node embedding initialization:} Following \citet{zhang2021greaselm}, we initialize the node embedding for the KG derived from UMLS and DrugBank using the pooled token representation of the node entity obtained from the SapBERT \cite{liu2021self}. To initialize the node embedding for the \textit{ConceptNet} KG, we utilized the approach proposed by \citet{feng2020scalable}, which converts each KG triplets into sentences that passed to the BERT-Large \cite{devlin2019bert} model to obtain the entity representation by applying the mean-pooling on entity mentions in the sentence.
\paragraph{Evaluation}:
Following the existing works on BioASQ8B, we evaluated the performance of the models using 
in terms of Mean Average Precision (MAP) \cite{tsatsaronis2015overview}, Recall@100 (R@100), and Normalised Cumulative Discount Gain (nDCG@10) \cite{jarvelin2002cumulated}. We use the official BioASQ script\footnote{\url{https://github.com/BioASQ/Evaluation-Measures}} to compute MAP values and used {Pytrec\_eval} \cite{van2018pytrec_eval} to report the nDCG@10 and Recall@100 score. Following \citet{thakur2021beir}, we report the Capped Recall@100 score for the TREC-COVID dataset.
\paragraph{Experiemental Setups:}
We utilized the pre-trained T5-base model from HuggingFace\footnote{\url{https://huggingface.co/t5-base}} \cite{wolf-etal-2020-transformers} to fine-tune it according to MonoT5 setup \cite{nogueira2020document} where we consider the query and gold document as a positive question-document pair and randomly taken the two other document from corpus which are not the part of the query's gold document to form the negative question-document pairs. We use Elasticsearch BM25 to report the lexical retrieval performance on all the datasets. In all our experiments, we re-rank the top 100 documents retrieved using BM25. For the BioASQ8B dataset, we use the $S=3$ and $R=9$, and the number of nodes in the subgraph is $10$. For the HotPotQA dataset, we use the $S=5$ and $R=7$, and the number of nodes in the subgraph is $15$. For both datasets, we find the optimal value of GNN hidden state representation size=$200$, the value of $\alpha=0.01$, and the projection dimension of the feed-forward network is $100$. The MonoT5 model is trained with batch size $16$, and GraphMonoT5 is trained with batch size $8$. We fine-tuned each model for $3$ epochs on BioASQ8B and HotPotQA datasets. The maximum token length of concatenated query and document is set to $512$ for all the experiments. The model parameters are updated using Adam \cite{kingma2015adam} optimization algorithm with the learning rate of $3e-4$ in all the experiments. We obtained the value of the optimal hyperparameters based on the respective development dataset performance in terms of nDCG@10 score.

\end{document}